\documentclass[final,5p,times,twocolumn,authoryear]{elsarticle}

\usepackage{amsmath,amsfonts,bm}

\def\eqref#1{equation~\ref{#1}}

\def\1{\bm{1}}

\def\vd{{\bm{d}}}

\def\vx{{\bm{x}}}
\def\vy{{\bm{y}}}
\def\vz{{\bm{z}}}

\def\mF{{\bm{F}}}

\def\mM{{\bm{M}}}

\def\mW{{\bm{W}}}

\DeclareMathAlphabet{\mathsfit}{\encodingdefault}{\sfdefault}{m}{sl}
\SetMathAlphabet{\mathsfit}{bold}{\encodingdefault}{\sfdefault}{bx}{n}

\usepackage{amsmath,amsfonts}
\usepackage{algorithmic}
\usepackage{algorithm}
\usepackage{array}
\usepackage[caption=false,font=normalsize,labelfont=sf,textfont=sf]{subfig}
\usepackage{textcomp}
\usepackage{stfloats}
\usepackage{hyperref}
\usepackage{url}
\usepackage{verbatim}
\usepackage{graphicx}
\usepackage{cite}

\usepackage{siunitx}

\usepackage{xspace}
\newcommand{\habitat}{\text{Habitat}\xspace}
\newcommand{\igibson}{\text{iGibson}\xspace}
\newcommand{\gibson}{\text{Gibson}\xspace}

\journal{}

\begin{document}

\begin{frontmatter}



\title{Insect-inspired Visual Point-goal Navigation} 

\author[ed,gla]{Yihe Lu} 
\author[ed]{Barbara Webb} 

\affiliation[ed]{organization={School of Informatics, University of Edinburgh},
            city={Edinburgh},
            state={Scotland},
            country={United Kingdom}}

\affiliation[gla]{organization={University of Glasgow},
            city={Glasgow},
            state={Scotland},
            country={United Kingdom}}

\begin{abstract}

Insect neuroethology provides a compelling biological template for efficient autonomous navigation.
We draw an analogy between the formal embodied AI visual point-goal navigation task and the ability of insects to discover, learn, and refine visually guided paths around obstacles between a discovered food location and their nest. 
We develop a novel integrative model of mushroom body and central complex, two insect brain structures, that have been implicated, respectively, in associative learning and path integration.
We demonstrate the mushroom body learning triggered by collisions results in adaptive obstacle avoidance and consequently optimised paths to the goal, 
corroborating the hypothesis of recent behavioural work that an insect can learn continuously as they travel.
The embodied insect-inspired model achieves success rates comparable to recent state-of-the-art models at many orders of magnitude less computational cost in the standardised Habitat point-goal navigation benchmark.
Testing in a more realistic simulated environment validates its robustness to perturbations.

\end{abstract}




\end{frontmatter}



\section{Introduction}
Goal-directed navigation requires an intelligent agent, artificial or biological, to reach a goal through motion planning and control. 
A particularly interesting form is to consider an agent aims to reach a predetermined position in a novel environment.
This is observed in the behaviour of insects,
such as bees, attempting to reach food sources found and communicated by other nestmates,
and has wide real-world applications in mobile robots, 
from indoor to outdoor, spanning ground, underground, marine, and aerial scenarios.
In an obstacle-free environment, reaching predetermined positions can be solved by Path Integration (PI), i.e., dead reckoning, solely.
In a more complex environment, however, 
an agent must employ other techniques to perceive obstacles (or traversable space) and take appropriate, ideally predictive, actions to find and follow an efficient path towards the goal.

This form of goal-direct navigation has been standardised, termed \textit{point-goal navigation}, and popularised by the Habitat navigation challenges
`to benchmark and accelerate progress in embodied artificial intelligence' \citep{savva2019habitat}.
In particular, at the start of point-goal navigation, the agent is provided with the explicit two-dimensional coordinates of the goal position (a point with a small catchment area)
and tasked to reach it without an \textit{a priori} map.
Typically, the Habitat point-goal navigation task considers an agent relying primarily on vision from a front-facing camera mounted on a robot.
Two contrasting solutions have been widely adopted over the years. 
A traditional approach is to create, update, and use a map on the fly,
i.e., perform visual Simultaneous Localisation And Mapping (SLAM).
As the map grows in size and fidelity, motion planning becomes less error-prone.
More recently, Reinforcement Learning (RL) approaches have increased in popularity, 
often making the assumption that there is no need for an explicit map.
In particular, an End-to-end RL model is expected to learn the direct correspondence between sensory inputs and (desired) control outputs,
using general-purpose RL methods.
Notably, the state-of-the-art agent,
reported to achieve \textit{perfect} performance in the Habitat point-goal navigation task \citep{ramakrishnan2021habitat},
was obtained by End-to-end RL.
More specifically, \citet{wijmans2019dd} developed a near-perfect agent by massively scaling up Proximal Policy Optimisation (PPO), the \textit{de facto} standard method of deep RL, into Decentralised Distributed PPO (DDPPO), and later \citet{ramakrishnan2021habitat} obtained the perfect agent with the same DDPPO method but using a larger and more accurately curated dataset of virtual scenes.

Here we adopt a third approach inspired by the homing behaviour of insects after foraging. 
It is a direct parallel to point-goal navigation,
as many insects are able to maintain, through PI, an estimate of their position relative to the nest over long journeys away from it, and will attempt to return on a straight line towards it (rather than the foraging journeys) when food is located \citep{muller1988path}. 
Moreover, during this return, the insect acquires visual information \citep{collett1992visual} that improves its efficiency in negotiating obstacles \citep{wystrach2011views}, leading to faster returns along improved paths on subsequent trips \citep{freas2025visual}.
Recent neuroethological and modelling studies have established plausible accounts of the neural circuits supporting these capabilities in insects,
with a focus on two brain structures that are conserved and anatomically stereotypical across insect species.
In particular, the \textit{Mushroom Body (MB)} is specialised in associative learning \citep{aso2014neuronal, webb2024beyond} and novelty (familiarity) detection \citep{hattori2017representations},
and the \textit{Central compleX (CX)} in PI and sensorimotor steering \citep{stone2017anatomically} (see \citep{webb2019internal} for a recent review on both structures).

Over the last decade, the CX has been clearly established as the `compass' circuit in the insect brain, combining multi-modal sensory inputs and self motion to maintain an estimate of heading direction with respect to an external reference frame and to control steering towards an internally specified goal direction \citep{hulse2021connectome, mussells2024converting, mitchell2026framework}. A minimal augmentation of this circuit provides a basis by which an insect can integrate its velocity over a foraging journey \citep{stone2017anatomically}, i.e., perform PI to estimate its own location and use this to move towards any position specified in the same reference coordinates \citep{le2019central}. Models of this circuit have shown its efficacy for such point-goal navigation 
\citep{goldschmidt2017neurocomputational,sun2020decentralised, goulard2023emergent,wystrach2023neurons},
including real-world tests on an aerial robot \citep{stankiewicz2020using}. However, this work has either neglected obstacles or added simple obstacle avoidance mechanisms to negotiate obstacles that prevent direct movement towards the goal \citep{baddeley2012model, sun2023insect}.

The insect's ability to memorise and follow visually familiar routes has been independently explained as deploying a `visual compass', which operates in a fundamentally different way. As first proposed by \citet{philippides2011might, baddeley2012model}, this involves memorisation of multiple panoramic views and the use of template matching \citep{labrosse2006visual}
to estimate view familiarity or novelty and thus align to the route direction.
\citet{ardin2016using} explicitly identified the circuit of the MB as a plausible substrate for this memory in the insect brain
and verified their spiking neural model in the simulated task of visual route following.
In addition to follow-up work on such MB models in route following or the standardised navigation task, `teach and repeat' \citep{kodzhabashev2015route, steinbeck2024familiarity, yihe2025insect}, the same idea of the MB being a learnable visual compass is also applied to or corroborated by work on visual homing,
similar to point-goal navigation,
but the same (non-novel) scene is used in testing as in training
and the agent can rely only on vision (no odometry) in testing \citep{moller2000insect, wystrach2020lateralised, zhong2025comparative}.

Several existing models integrate the CX and the MB.
The agents of \citet{baddeley2012model, amin2025ant} perform homing relying only on (non-visual) PI and obstacle avoidance in its first attempt
and switch to visual route following in subsequent trials.
\citet{sun2020decentralised, goulard2023emergent} have explored more organic integration of the two circuits,
demonstrating emergent behaviours, e.g., automatic switch from PI to visual homing when the embodied model is displaced to an arbitrary position, maintaining goal approaching locomotion when target object is temporarily obscured or removed.
However, all these integrative models have an initial phase dedicated to training, 
implicitly assuming the optimality of the training route(s), 
and, consequently, do not address how such routes are \textit{formed} at the first place 
or \textit{refined} after the initial training phase. 
These questions have been thoroughly studied in behavioural experiments \citep{freas2017learning, freas2019terrestrial, clement2024latent, freas2025visual}. 
In particular, \citet{clement2024latent} explicitly demonstrated `ants learn \textit{continuously} the routes they travel' and provided a high-level model of how this might occur, but without addressing the neural implementation.

Here, we aim to develop a novel integrated model of the MB and the CX and assess the performance in realistic indoor robotic point-goal navigation simulations. 
In brief, the CX is responsible for determining control signals based on desired (mostly goal-directed) and actual orientations,
while the MB learns the association between views and obstacles \textit{on the fly}
and uses the learnt association to modulate the CX computation simultaneously.
We hypothesise that, starting with zero prior memory,
our model can gradually learn to avoid obstacles in a predictive manner,
improving its path towards a goal even on the first trial, but increasingly so after repetition, 
which may even lead to a successful trial after failures.
We first test our model in the \habitat simulator \citep{savva2019habitat} (offering the standardised point-goal navigation task of the 2019 Habitat Challenge),
permitting direct performance comparison between our model and others reported in recent literature.
We then deploy our model in the \igibson simulator \citep{li-iGibson}, which offers a more realistic physics engine,
to validate its sensorimotor robustness by adding artificial noise and bias.
In both simulators, we examine the contributions from different pathways of the model through ablation studies.

This paper is organised as follows:
\S\ref{sec-method:our-model} introduces our insect-inspired model,
\S\ref{sec-method:setup} describes the experimental setup,
\S\ref{sec-result} presents the main experiment results,
and \S\ref{sec-discussion} discusses the implications.

\section{Methods}
\label{sec-method}

\subsection{Insect-inspired Model}
\label{sec-method:our-model}

\begin{figure}[t!]
    \centering
    \includegraphics[width=0.48\textwidth]{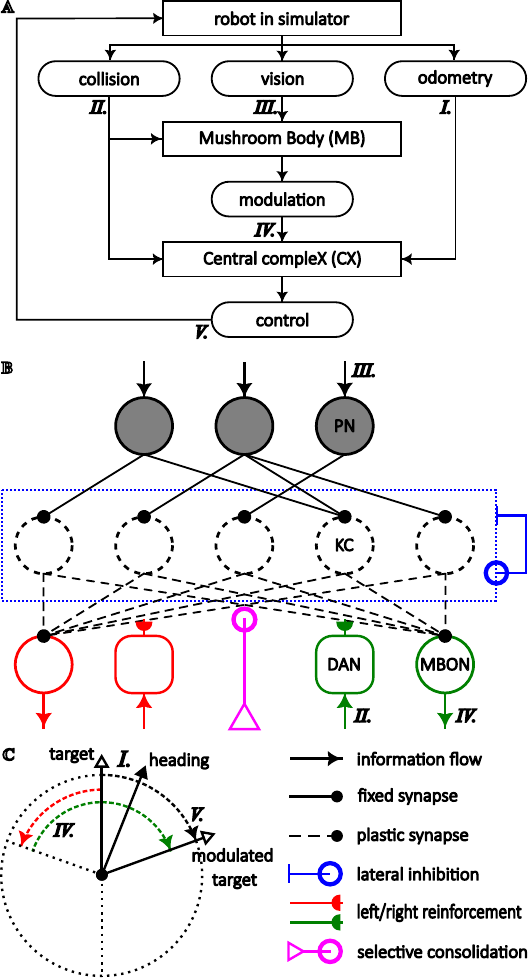}
    \caption{Overview of our insect-inspired model.
    \textbf{(A)} The closed-loop learning and control of the model on a virtual robot. 
    The model is primarily composed of the MB and CX modules, using three types of (preprocessed) data as inputs, odometry (\textbf{\textit{I}}), collision (\textbf{\textit{II}}), and vision (\textbf{\textit{III}}) to determine control commands of the robot (\textbf{\textit{V}}).
    \textbf{(B)} The neural architecture of the MB.
    The PN activity encodes preprocessed visual information (\textbf{\textit{I}}),
    and the MBON activity becomes the modulatory signal of the MB on the CX (\textbf{\textit{IV}}).
    The PN-KC synapses are randomly fixed, binary, and sparse.
    Only the KC-MBON synapses are plastic,
    and learning is gated by the DANs, encoding collision-related signals (\textbf{\textit{II}}).
    Only if task performance is improved, recent memories are consolidated.
    \textbf{(C)} The angular calculation performed by the CX. A target direction is set, either towards the goal or to escape an obstacle, and then modulated by the MB signals (\textbf{\textit{IV}}).
    The modulated target is then compared to the current orientation (\textbf{\textit{I}}) to calculate the desired rotation.
}
    \label{fig:model-overview}
\end{figure}

Our insect-inspired model is primarily composed of two modules, the CX and the MB, emulating in simplified form the corresponding insect brain structures and their functions (Fig. \ref{fig:model-overview}A).
The CX generates control commands and performs PI using odometry inputs.
By default, it chooses the direction towards the goal as the desired target direction.
After a collision, it sets a temporary target direction different from the goal direction,
so that the model attempts to escape from the obstacle.
In addition, the MB is triggered to learn visual inputs just before collisions and after escapes.
Based on such collision-related visual memory, the MB modulates the target direction chosen by the CX constantly,
which is expected to reduce risks of future collisions by recognising visually similar obstacles.

\subsubsection{Visual Preprocessing}
\label{sec-method:visual-preprocessing}
The raw RGB images from the front-facing camera of the robot at time \(t\) are greyscaled, 
filtered by difference of Gaussians for feature enhancement \citep{marr1980theory} (potentially edge enhancement, cf. Sobel filter in \citet{gattaux2025route}),
downsampled to \(33\times 33\) pixels, and then flattened into an array, \(\vx_t\).
Such low-resolution vision is consistent with insect compound eye optics \citep{schwarz2011properties, seung2024predicting} 
and proved sufficient to produce good (or even better) performance while reducing computational costs \citep{wystrach2016field, steinbeck2024familiarity, gattaux2025route, yihe2025insect}.

\subsubsection{Associative Learning}
\label{sec-method:MB}
Learning is performed by the MB, an artificial neural network consisting of \(N_\text{PN}=1089\) Projection Neurons (PNs), \(N_\text{KC}=32000\) Kenyon Cells (KCs), 
\(2\) MB Output Neurons (MBONs), and \(2\) DopAminergic Neurons (DANs) (Fig. \ref{fig:model-overview}B).
The neural architecture is primarily feedforward from PNs to KCs to MBONs with the PN-KC synaptic weight matrix denoted by \(\mM\in\{0,1\}^{N_\text{KC} \times N_\text{PN}}\) and the KC-MBON matrix by \(\mW_t\in[0,1]^{2 \times N_\text{KC}}\).
The PN activation encodes the preprocessed visual input, \(\vx_t\in[0,1]^{N_{\text{PN}}}\),
which is transformed into a latent KC representation by
\begin{equation}
\label{eq:pn-kc}
    \vy_t = g(\mM \vx_t),
\end{equation}
where \(g(\cdot)\) is the \(k\)-winners-take-all function with \(k=320\), 
emulating lateral inhibition amongst real KCs that results in sparse \(\vy_t\in\{0,1\}^{N_\text{KC}}\).
Subsequently, the two, left and right, MBON outputs, \(\vz_t = [z_{\text{L}, t}, z_{\text{R}, t}]^\top \in [0,1]^2\), are obtained by
\begin{equation}
\label{eq:kc-mbon}
    \vz_t = \frac{1}{k}\mW_t \vy_t.
\end{equation}

We highlight that \(\mW_t=[\mW_{\text{L}, t}, \mW_{\text{R}, t}]^\top\) is the \textit{only} plastic component within the MB and the entire model,
whereas \(\mM\) is fixed (thus not indexed by \(t\)).
In particular, \(\mM\) is randomly initialised to be sparse and binary as in a real MB by \(\mM \sim \text{Bernoulli}(N_\text{PN/KC}/N_\text{PN})^{N_\text{KC} \times N_\text{PN}}\), where \(N_\text{PN/KC}=10\) denotes the average number of PNs connected to one KC.
In contrast, \(\mW_0=\mathbf{1}_{2\times N_\text{KC}}\) is initialised to be all-to-all,
and the plasticity of \(\mW_t\) obeys the hetero-associative learning rule
\begin{equation}
\label{eq:learn-depression}
    \Delta\mW_t = -\alpha \vd_t \vy^\top_{t-\tau_d} \odot \mW_{t-\tau_d},
\end{equation}
where \(\alpha=1\) is the learning rate,
\(\vd_t=[d_{\text{L}, t}, d_{\text{R}, t}]^\top\in\{0,1\}^2\) the DopAminergic Neuron (DAN) activities,
and \(\tau_d\) the time gap between visual and collision-related signals.
\(\vd_t\) is responsible for gating, as the left and right DANs encode reinforcement signals \citep{aso2014neuronal}
and are assumed to be activated by collision-related events here.
When the robot collides with an obstacle on its left or right side,
the ipsilateral DAN is activated for receiving a punishment.
When the robot is released from escape induced by the collision, the contralateral DAN will be activated, 
because removal of punishment is considered rewarding \citep{tanimoto2004event}.

Functionally, the MB's neural architecture with a single MBON approximates a Bloom filter for novelty detection \citep{dasgupta2018neural}.
As \(\mW_t\) is updated only according to visual inputs just before collisions and after escapes,
\(z_{\text{L}, t}, z_{\text{R}, t}\) encode effectively the left and right visual novelty of obstacles,
i.e., how the model perceives the absence of obstacles,
and the plasticity of \(\mW_t\) results in adaptive collision avoidance (\S\ref{sec-model:cx}, cf. the more abstract model in \citet{porr2003isotropic}).

\subsubsection{Memory Consolidation}
\label{sec-model:memory-consolidation}
Inspired by the different time scales and selective memory consolidation of real MB memory  \citep{davis2011traces, gkanias2022incentive},
we partition the learnt memory (total change in weights) into Short-Term Memory (STM), Intermediate-Term Memory (ITM), and Long-Term Memory (LTM), i.e.,
\begin{equation}
    \mW_0 - \mW_t = \Delta\mW^\text{LTM}_k + \Delta\mW^\text{ITM}_k + \Delta\mW^\text{STM}_k,
\end{equation}
according to the temporal hierarchy of the navigation task (\S\ref{sec-method:setup}).
Specifically, the model is tested throughout a sequence of trials.
Before the first trial, \(\Delta\mW^\text{STM}_1 = \Delta\mW^\text{ITM}_1 = \Delta\mW^\text{LTM}_1 = \mathbf{0}\) as \(\mW_0\) is initialised. 
In every trial \(k\), \(\Delta\mW^\text{STM}_k = \sum_{t\in\mathcal{T}_k}\Delta\mW_t\), where \(\mathcal{T}_k\) denotes the time interval of the current trial.
When the trial ends, selective memory consolidation is performed by
\begin{subequations}
\begin{align}
    \Delta\mW^\text{LTM}_{k+1} &= \Delta\mW^\text{LTM}_k + \mathbb{I}\left(\text{SPL}_k \geq \text{SPL}_{k-1}\right) \Delta\mW^\text{ITM}_k, \label{eq:ltm} \\
    \Delta\mW^\text{ITM}_{k+1} &= \Delta\mW^\text{STM}_k, \\
    \Delta\mW^\text{STM}_{k+1} &= \mathbf{0}, \label{eq:stm}
\end{align}
\end{subequations}
where SPL is a performance metric defined in \S\ref{sec-methods:metric}.
Thus, only if recent memory benefits performance, ITM is consolidated into LTM.

\subsubsection{MB-modulated Steering}
\label{sec-model:cx}
The CX is not implemented as a neural network,
as it is not plastic in our model.
Nevertheless, it performs a central role of sensorimotor control
and is abstracted to be a module calculates the desired rotational angle as
\begin{equation}
\label{eq:angle2rotate}
    \Delta\sigma_t = \theta_t - \sigma_t + \phi_t,
\end{equation}
where \(\theta_t\) represents the target direction, 
\(\sigma_t\) the agent's orientation,
and 
\begin{equation}
\label{eq:mb-modulation}
    \phi_t = \pi (z_{\text{L}, t} - z_{\text{R}, t}),
\end{equation}
the MB modulation (Fig. \ref{fig:model-overview}C).
\(\theta_t\) is chosen to be the direction to the goal, \(\theta^*_t\), if no collisions have occurred recently for \(\tau_c\), the duration of escape.
After colliding with an obstacle at \(t_c\), a temporary target direction, \(\theta'_t\), is immediately set to the collision direction, \(\gamma_c\) (\S\ref{sec-method:sensory-input}), and gradually changes towards (not necessarily reach) the tangent direction, \(\gamma_e\), satisfying \(\gamma_e \perp \gamma_c\) and \(|\gamma_e - \sigma_{t_c}|< \pi/2\), for \(t\in[t_c, t_c + \tau_c]\).
All the angular variables are in radians.
In particular, \(\phi_t \in [-\pi, \pi]\) as \(z_\text{L}, z_\text{R} \in [0, 1]\).

Starting with \(\mW_0=\mathbf{1}_{2\times N_\text{KC}}\), the agent attempts to move towards the goal in a straight line as \(\phi_t = 0\), until the robot collides with an obstacle.
If the collision occurs on the left side, \(\mW_{\text{L}, t}\) will be depressed.
As a result, \(z_{\text{L},t} < z_{\text{R},t}\) is expected in future when the agent is facing the same or a similar obstacle,
i.e., the right side is more attractive,
and consequently the MB modulation will promote a right turn.

The abstracted angular computation by Eq. (\ref{eq:angle2rotate}) is viable,
as the real CX is shown to track body orientation and goal direction \citep{stone2017anatomically, hulse2021connectome},
to compare the two to generate goal-directed steering signals \citep{mussells2024converting, mitchell2026framework},
and to be innervated and modulated by the MB \citep{aso2014neuronal,li2020connectome}.
Calculating \(\phi_t\) by Eq. (\ref{eq:mb-modulation}) is similar to tropotaxis,
the renowned locomotion strategy using lateral sensory difference,
which has been shown to be viable and effective in recent work of visual route following by insect MB models \citep{wystrach2020lateralised,wystrach2023neurons,steinbeck2024familiarity,yihe2025insect}.
\\

More details of the user-defined variables, e.g., \(\tau_d, \tau_c\) can be found in Table S1.

\subsection{Experiment Setup}
\label{sec-method:setup}
Generally in point-goal navigation, an agent is expected to drive a robot from a start position to a goal position using only its sensors,
given no map (information of obstacles) but only the two-dimensional coordinates of the two positions.
This work, in particular, considers agents using RGB inputs from the front camera, 
accurate odometry (position and orientation, i.e., GPS+Compass) of the robot,
and information of collisions between the robot and obstacles.

We tested our agents in independent \textit{episodes},
each composed of a sequence of \(20\) \textit{trials}.
A trial would terminate after \(500\) frames or earlier if the agent reached the goal successfully (\S\ref{sec-methods:metric}).
For any trial of the same episode, the scene, start position and orientation, and goal position are identical.
Throughout the trials of an episode, the MB would keep learning (\S\ref{sec-model:memory-consolidation}), 
and consequently its modulation on the orientation would grow stronger until no collision occurred or the MB memory saturated.
Across episodes, the memory was completely reset.

\subsubsection{Simulation Platforms}
\label{sec-preliminary:test-platform}

We used two simulators, \habitat \citep{savva2019habitat} and \igibson \citep{li-iGibson}, 
with default settings.
Both simulators offered realistic visual rendering, with the main limit on realism that the lighting was perfectly consistent.
Kinematics in both simulators were deterministic, except when artificial noise was added.
In addition, robots and other objects were perfect rigid bodies.
They were different primarily in physics.
While all objects except the robot in \habitat were immovable,
some objects in \igibson objects were movable by the robot, e.g., chairs.
Moreover, \igibson employed a realistic physics engine,
simulating object motions based on the classical kinetics,
whereas \habitat assumed significantly simplified kinematics,
discretising robot motions completely,
so that large-scale (RL) simulations could be performed more efficiently.

Our model were deployed on a primitive cylindrical `robot' in \habitat
and the Freight robot in \igibson (scaled down from the default size by a factor of \(0.3\)),
both equipped with a front-facing RGB camera (see Table \ref{tab:robots} for the specifications).
Freight was chosen amongst the robots available in \igibson, 
for its shape the most comparable to the cylindrical robot in \habitat and its maximal speed capacity the highest,
implying the longest possible distance of movement or a potentially minimal testing duration but a maximal difficulty for steering due to inertia (\S\ref{sec-model:motion-control}).

\begin{table}[t!]
    \centering
\caption{Robot specifications.}
\label{tab:robots}
    \begin{tabular}{ccc}\hline  
        Simulator         & \igibson                        & \habitat\\ \hline 
        Footprint Diameter& \qty{0.1677}{\meter}            & \qty{0.2}{\meter}\\
        Camera Height     & \qty{0.0594}{\meter}            &\qty{1.5}{\meter}\\
        \(v_{\max}\)      & \qty{2.133}{\meter\per\second}  & \qty{0.25}{\meter} frame\(^{-1}\)\\
        \(\omega_{\max}\) & \qty{11.469}{\radian\per\second}& \ang{10}  frame\(^{-1}\)\\ 
        Field of View     & \ang{90} \(\times\) \ang{90}    & \ang{90} \(\times\) \ang{90}\\ \hline 
    \end{tabular}
    
\end{table}


We used both simulators with virtual reconstructions of real-world indoor scenes from the \gibson dataset \citep{xiazamirhe2018gibsonenv}.
\habitat was used with the \(14\) validation-split scenes from a larger set of high quality reconstructed scenes, known as \gibson 4+, manually curated and rated by \citet{savva2019habitat},
and \igibson with its \(15\) default scenes.
In addition, \(93\) standardised \habitat episodes were used for model comparison (\S\ref{sec-method:sota}),
while \(300\) \igibson episodes were randomly generated.

\subsubsection{Sensory Inputs}
\label{sec-method:sensory-input}
The field of views of the front-facing RGB cameras on the two robots were identical (Table \ref{tab:robots}).
Considering additionally that (at least part of) the same \gibson dataset were used,
visual inputs in the two simulators would be similar in terms of generic visual features.
A major difference was expected to be derived from the robots' different heights.
As the camera was placed at a lower height on Freight,
visual inputs in \igibson always contained a larger portion of floor than in \habitat.

An RGB, rather than RGBD, camera was chosen,
primarily because RGB cameras are more widely applicable.
While the extra depth sensing could always bring performance improvement in simulation \citep{mishkin2019benchmarking, savva2019habitat, wijmans2019dd, zhao2021surprising, partsey2022mapping, ramakrishnan2021habitat},
the magnitude of such improvement is largely variable. 
All others equal, a model with depth-only sensing could outperform the same model with an RGB camera \citep{mishkin2019benchmarking, ramakrishnan2021habitat}
and even an RGBD camera \citep{savva2019habitat}.
Note our model could have used a monochrome camera for greater compatibility with insect vision and potentially enhanced efficiency in its visual preprocessing (\S\ref{sec-method:visual-preprocessing}).
However, RGB camera is necessary for direct comparison with existing models.

The position and orientation data were extracted directly from the simulators
and thus perfectly accurate (with the default settings).
Although this assumption is unrealistic, 
insects can robustly estimate odometry and perform PI by sensory cue integration, primarily using the CX \citep{webb2019internal, noorman2024maintaining},
and \citet{goldschmidt2017neurocomputational} demonstrates in simulation the noise robustness of a CX (odometry)-only odel.
Moreover, the assumption is consistent with other recent point-goal navigation work \citep{mishkin2019benchmarking, savva2019habitat, wijmans2019dd, zhao2021surprising, partsey2022mapping, ramakrishnan2021habitat}.

The information of collision direction, \(\gamma_c\),  necessary for the model (\S\ref{sec-model:cx}), was not extracted directly from the simulators.
In \habitat, after a collision event recorded by the simulator,
\(\gamma_c\) was approximated by the direction of the vector from the anticipated position (given the last action) to the actual position (\S\ref{sec-model:motion-control}).
In \igibson, the magnitude of the net horizontal force on the robot, \(\mF_h\), was compared to a threshold, \(F_c=\qty{100}{\newton}\).
If \(|\mF_h| \geq F_c\), a collision event was recorded with \(\gamma_c\) set to be the direction of \(\mF_h\).

\subsubsection{Motion Control}
\label{sec-model:motion-control}
Due to the drastic difference between the underlying physics of the two simulators,
the permitted control signals were completely different.
In \habitat, only one action at each frame, \(a_t\), can be chosen out of the three options:
move-forward (by \qty{0.25}{\meter}), turn-left (rotation by \ang{10}), and turn-right (rotation by \ang{-10}).
Inertia and force (acceleration) were not considered.
Given any control signal, the robot would change its position or orientation instantaneously within the frame.
This change was perfect, unless a collision occurred,
in which case the robot would slide against the obstacle with a displacement determined by geometry \citep{savva2019habitat}.

In contrast, the realistic physics engine of \igibson approximates continuous physics at the default simulation frequency of \qty{120}{\hertz},
and the robot was controlled according to a pair of normalised linear and angular speeds, \(v_t, \omega_t \in [-1, 1]\),
at the default visual rendering frequency of \qty{30}{\hertz}. 
The unnormalised speeds would then be obtained by multiplying \(v_{\max}, \omega_{\max}\) (Table \ref{tab:robots}).
Unlike in \habitat, however, these speeds as control signals could not be reached instantaneously,
as the actuators were also physically realistic.

Consequently, after computing \(\Delta\sigma_t\) by Eq. (\ref{eq:angle2rotate}), the low-level motion control of the robots had to be different for our model in these two simulators.
In \habitat, \(a_t\) was set to be turn-left or turn-right if \(|\Delta\sigma_t| \geq \ang{10}\);
otherwise, \(a_t\) was set to be move-forward.
In \igibson, \((v_t, \omega_t) = (\cos \Delta\sigma_t, -\sin \Delta\sigma_t)\).

For determinism, we further assumed that an embodied model could complete all necessary computations from sensory input processing to control signal generation within each frame.

\subsubsection{Performance Metrics}
\label{sec-methods:metric}
We considered three metrics for measuring performance in the point-goal navigation task:
\begin{itemize}
    \item Success Rate (SR): whether the agent could reach the goal position with a catchment area of \qty{0.2}{\meter} radius,
    \item Success weighted by Path Length (SPL):
        \begin{equation}
        \label{eq:spl}
            \text{SPL} = \text{SR} \cdot \frac{l}{p},
        \end{equation}
        where \(l\) denotes the geodesic distance, i.e., the length of the shortest path, between the start and the goal positions and \(p\) the length of the path taken by the robot under autonomous control,
    \item Total number of collisions.
\end{itemize}
Note by definition \(p\geq l\) and thus \(\text{SPL} \leq \text{SR}\) is always true.
SPL and SR are widely reported in the recent literature of point-goal navigation.
The third metric was important for demonstrating that our agent, which used collision information (\S\ref{sec-model:cx}),
could not just exploit passive sliding against obstacles as other agents have been shown to do in these simulation environments \citep{kadian2020sim2real}.

\subsection{Model Comparison}
\subsubsection{State-of-the-art Results}
\label{sec-method:sota}
For benchmarking purpose, we identified seven models (amongst numerous variations) reported in six recent works on point-goal navigation \citep{mishkin2019benchmarking, savva2019habitat, wijmans2019dd, zhao2021surprising, partsey2022mapping, ramakrishnan2021habitat}.
All the works tested their models in \habitat with the standardised episodes in the \gibson 4+ scenes \citep{savva2019habitat},  
except \citet{mishkin2019benchmarking}, published before the Habitat challenge \citep{savva2019habitat}, using the MINOS simulator with Matterport3D (MP3D) scenes \citep{savva2017minos}.

In addition, their experiment setups were identical to ours (\S\ref{sec-method:setup}),
except for four technical differences. 
\begin{enumerate}
    \item The standardised task did not provide models with collision data (at least in testing), whereas our task did.
    \item Visual inputs were noisy in \citet{zhao2021surprising, partsey2022mapping}, and their models had access to odometry data only in training but not in testing.
    \item There was only one trial per episode in testing of the standardised task.
    \item Our experiment did not contain a separate training phase.
\end{enumerate}
The provision of collision data did not give our model an edge in terms of information gain from sensory inputs, 
as it was computed from odometry data in \habitat (\S\ref{sec-method:sensory-input}), 
and the ability to detect collisions would be in any case a realistic assumption for robots and insects.
Similarly, the absence of visual noise was not considered critical,
because our previous work on visual route following \citep{yihe2025insect} has validated the visual robustness of a more primitive but similar MB model in extensive simulations and the real world. 
We discuss further in \S\ref{sec-discussion:mapless-nav} the arbitrary and implicit role of odometry in previous experiments, again noting that the availability of approximate odometry is a reasonable assumption for robots and insects.
As regards the third and fourth differences, our model performed learning in testing on the fly by design,
and it would be fair to compare our model's performance of first trials in episodes to these pre-trained models' test performance in novel scenes.

The seven models we use for comparison can be classified into three types (Table \ref{tab:model-types}), based on their different assumptions of the spatial prior:
\begin{itemize}
    \item \textit{End-to-end RL} assumes no spatial prior.
    An agent should learn to select actions based directly on sensory inputs.
    \item \textit{Visual Odometry and RL (VO+RL)} asserts the efficacy of odometry.
    An agent should learn how to locate and orient itself with respect to an external reference frame (VO)
    and how to select actions based on its position and orientation (RL).
    \item \textit{SLAM} relies on maps (or mapping).
    An agent should learn about its surroundings and then plan and perform its locomotion accordingly.
\end{itemize}

\begin{table}[t!]
    \centering
    \caption{Seven recent point-goal navigation models for comparison.
    More details are summarised in Table S2.  
    }
    \begin{tabular}{lllll}\hline
        Type           &Model &Reference \\ \hline 
        End-to-end RL  &E1    &Table 2 \citep{savva2019habitat} \\  
                       &E2    &Table 2 \citep{wijmans2019dd} \\
                       &E3    &Table 2 \citep{ramakrishnan2021habitat}\\ \hline
        VO+RL          &V1    &Table 2 \citep{zhao2021surprising} \\
                       &V2    &Table 1 \citep{partsey2022mapping} \\   \hline
        SLAM           &S1    &Table 3 \& 11 \citep{mishkin2019benchmarking} \\ 
                       &S2    &Table 2 \citep{savva2019habitat} \\ \hline
    \end{tabular}
    \label{tab:model-types}
\end{table}

All these models required extensive training before testing.
In particular, with billions of training frames, \citet{wijmans2019dd, ramakrishnan2021habitat, partsey2022mapping} trained their state-of-the-art models of end-to-end RL and VO+RL, achieving near-perfect or even perfect performance with RGBD inputs.
Note, for comparison, we take the results of the best RGB agent trained in Gibson scenes from each work (not their best agents),
except for V2 and S2 as only the RGBD performance is reported.

\subsubsection{Ablation Studies}
To examine the contributions of the three neural pathways, odometry, collision, and vision, of our model (`I', `II', and `III' in Fig. \ref{fig:model-overview}A),
we conducted ablation studies by removing firstly the vision pathway (producing the `odometry-collision model') and subsequently the collision pathway (producing the `odometry-only model', called `Goal-follower' in \citet{savva2019habitat, mishkin2019benchmarking, wijmans2019dd}). 
Note the ablation of the collision pathway only would effectively disable the vision pathway, as the MB learning had to be triggered by collisions,
making the potential `odometry-vision model' behaviourally equivalent to the odometry-only model.
Further, the ablation of the odometry pathway only would produce random exploration, similar to the `Random' model reported in \citet{savva2019habitat}, 
which yielded performance as poor as \(\text{SR}=0.03\).
Hence the odometry-only (A1) and odometry-collision (A2) models are the only relevant ablations to test. 

\section{Results}
\label{sec-result}

\subsection{Goal Reaching with Minimal Learning}
\label{sec-result:goal-reaching}

\begin{figure}[t!]
    \centering
    \includegraphics[width=0.48\textwidth]{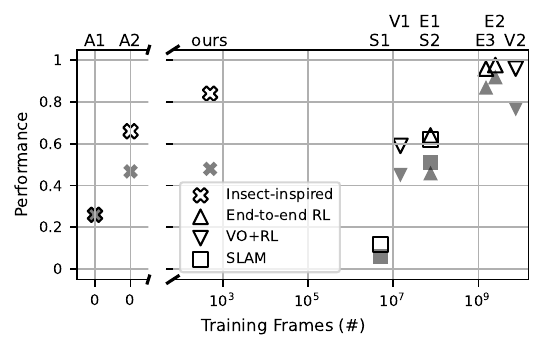}
    \caption{Model performance measured in SR (hollow) and SPL (filled) (defined in \S\ref{sec-methods:metric}) vs total number of simulation frames available for training.
    The results of our insect-inspired model and the ablated models (A1: odometry-only, A2: odometry-collision) were obtained from 100 independent test episodes in \habitat with the \gibson 4+ scenes.
    The results of the other models, cited in Table \ref{tab:model-types}, are taken from the literature (summarised in Table S2).
}
    \label{fig:benchmark}
\end{figure}

To highlight the efficacy and lightweightness of our insect-inspired model,
we compare the average performance of the model's first trials across \(100\) \habitat episodes (effectively using STM only)
with the results of the seven models in Table \ref{tab:model-types}.
While our model was outperformed by V2, E2, and E3,
it achieved better or at least comparable performance to S1, S2, V1, and E1, particularly in SR (Fig. \ref{fig:benchmark}).
The three best models required billions of training frames and the other four tens of millions,
whereas our model required no training before testing and used minimal online learning of up to \(500\) frames,
i.e., the duration of the first trial of an episode.

Minimal learning was effective for our model to reach goals successfully (\(\text{SR}= 0.84\)),
because the model relied on the strong prior of goal approaching,
and its learning focused on mitigating the potential failures of the prior (i.e., avoiding obstacles).
However, the prior helped the model little with optimising paths (\(\text{SPL}= 0.48\)),
because the model's success of reaching goals was obtained by trial and error.
It held absolutely no knowledge of the environment before its first trial in each episode
and had to learn to move around obstacles from collisions on the fly.
In contrast, when starting the testing, the RL and SLAM models would be able to visually recognise obstacles to some extent 
and could have even acquired some `common knowledge' of floor plans from training,
potentially benefitting their path planning in the test scenes.

Remarkably, despite A2 being unable to see or learn, its performance was comparable to that of S2, V1, and E1. The substantial improvement over A1 (\(\Delta\text{SR}=0.40\) and \(\Delta\text{SPL}=0.20\))
implies the importance of the collision pathway for the robot not getting `stuck' amongst obstacles.
While the additional capability of learning yielded negligible improvement in path optimality (\(\Delta\text{SPL}<0.01\)),
it enabled the full model to solve many episodes in which A2 failed (\(\Delta\text{SR}=0.18\)).

\subsection{Path Optimisation with Online Learning}
\label{sec-result:path-optimisation}

While it was challenging for the insect-inspired model to find an optimal path to a goal on the first trials,
it could potentially improve its performance in successive trials.
Note the end of learning would occur either after the maximum \(20\) trials or earlier once its LTM did not update after a trial, i.e., SPL dropped.
Subsequent trials were no longer necessary, 
because the model (MB memory) was reset to an earlier state
and it would have merely repeated the path (indeed every movement) of the second-to-last trial in the perfectly deterministic simulations.
For the same reason, the ablated models incapable of learning were tested only once per episode.

\begin{figure*}[t!]
    \includegraphics[width=\textwidth]{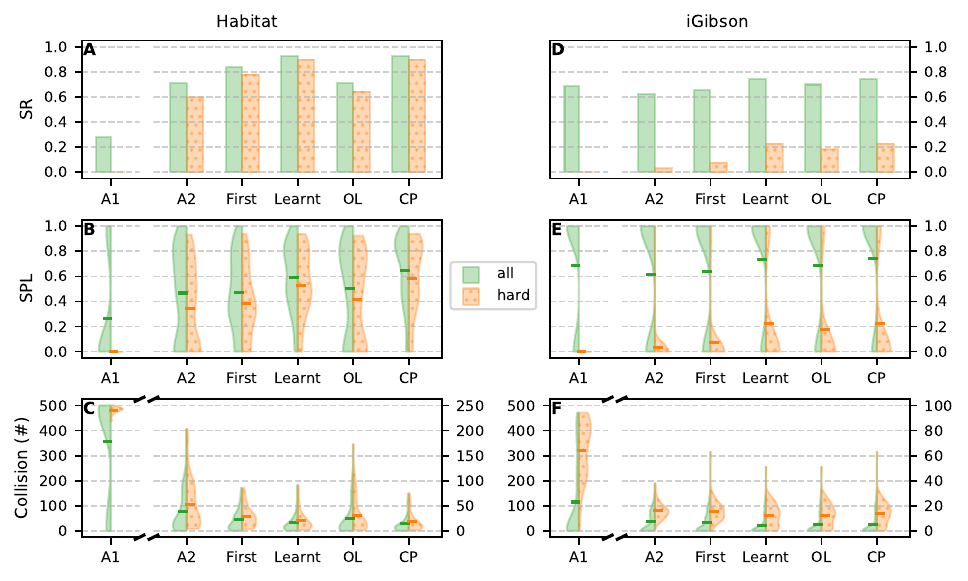}   
    \caption{Performance of different variants of our insect-inspired model in \habitat and \igibson.
    A1: the odometry-only model.
    A2: the odometry-collision model.
    First: the full model in the first trials.
    Learnt: the full model when it ended learning (i.e., detecting SPL drop, occurring within on average \(5\) trials).
    OL: the model performing unconditional memory consolidation and overloaded with all memory within an episode.
    CP: the model storing all episode experiences as checkpoints and cherrypicking the best performance.
    The episodes could not be solved by A1 were considered `hard', and thus the model scored \(0\) in \textbf{ABDE}.
    Note different scales are used within \textbf{CF}.
}
    \label{fig:model-performance}
\end{figure*}

In both simulators, the capability of online learning enabled the full model to consistently and quickly improve its performance from its first attempt until its end of learning (`First' and `Learnt' in Fig. \ref{fig:model-performance}). 
Performing PI only, A1 was unable to solve a subset of episodes (\(\text{SR}=0\)), considered `hard'.
We also plot the results for the other models on these `hard' episodes only,
which reveals a stronger effect of learning over repeated trials for the full model.

Interestingly, the more ablated A1 outperformed A2 in \igibson (Fig. \ref{fig:model-performance}DE).
We note that this result was obtained in \igibson but not in \habitat,
primarily because \igibson contained movable obstacles.
While A1 could push through them by brutal force,
A2 would reflexively escape.
In addition, in both simulators, any success of A1 was achieved by accruing a large number of collisions (Fig. \ref{fig:model-performance}CF),
because the robots, modelled as a perfect rigid body, could keep sliding against obstacles without any physical damage.
Generally in practice, this assumption is unrealistic for real robots, and (excessive) collisions are unwanted.
In contrast, the collision pathway, relying on the simple reflexive escape mechanism, brought the number of collisions down by an order of magnitude,
while learning reduced the collision risk further, as expected.

\subsection{Sustaining Performance by Memory Consolidation} 
\label{sec-results:memory-consolidation}

The selective memory consolidation mechanism (\S\ref{sec-model:memory-consolidation}) performed a critical role in sustaining the relative high performance after learning.
The model yielded worse performance on average, 
When consolidating all ITM into LTM unconditionally (`OL' in Fig. \ref{fig:model-performance}),
with memory overloaded in the MB
(\S\ref{sec-discussion:memory}),
the MB would grow more susceptible to memory interference, mistakenly anticipating the presence of an obstacle while facing traversable space. 
Consequently, the model would become over-cautious and more likely to get `stuck' in a cluttered area or diverted towards a significantly wrong direction.

While selective consolidation of ITM into LTM by Eq. (\ref{eq:ltm}) prevented performance drop,
it eliminated the possibility that the performance could rebound after a drop, i.e., the SPL might get stuck in a local minima, in the perfectly deterministic simulations.
Indeed, we could construct a hypothetical model that achieved a higher performance.
Specifically, it saved all data of weight updates (without selective consolidation) and the SPL at the end of every trial as checkpoints and 
simply cherry-picked for each episode the checkpoint with the highest SPL (`CP' in Fig. \ref{fig:model-performance}). 

Although the selective memory consolidation mechanism of our model would not prevent SPL rebound in non-deterministic environments (stochastic simulations in \S\ref{sec-result:robust-control} or any realistic environments),
and the hypothetical model was impractical for both insects and robots considering its small performance improvement at great memory cost,
it prompts consideration of alternative memory consolidation mechanisms (\S\ref{sec-discussion:memory}).

\subsection{Robust Control Using Reafference Only}
\label{sec-result:robust-control}

Our model was finally validated in \igibson with imperfect motor control.
In particular, the angular speed was constantly perturbed by artificial steering noise, \(n_\omega \sim \text{Lognormal}(0, \sigma_\omega^2)\), and bias, \(b_\omega\in(-1, 1)\), that is,
\begin{equation}
\label{eq:steer-perturb}
    \omega_\text{motor} = \text{Clip}((\omega_\text{model} + b_\omega) \cdot n_\omega, -1, 1).
\end{equation}
As the noise eliminated perfect determinism from our simulations, 
it would be possible for the model to learn and behave differently in subsequent trials after a performance drop,
and thus we would not stop an episode earlier than \(20\) trials as in the previous sections.

We emphasise that the model controlled the robot using visual reafferent inputs only.
The perturbation by Eq. (\ref{eq:steer-perturb}) was unknown to the model,
which received no efference copy by design.
Even if the model could be potentially re-designed to process an efference copy,
the artificial perturbation could still emulate imperfections in the robot's actuation and unknown factors in the environment.
For comparison, A2 was tested under identical conditions to the full model.

\begin{figure}[ht!]
    \centering
    \includegraphics[width=0.48\textwidth]{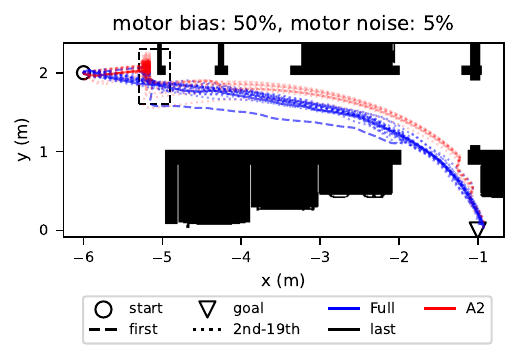}
    \caption{An example experiment with strong motor perturbation, \(b_\omega = 0.5\) and \(n_\omega=0.05\).
    In \(20\) consecutive trials, the robot started at \((-6, 2)\) (circle with size approximating the footprint of the robot) and the goal was to reach \((-1, 0)\) (triangle) within \(500\) frames.
    All the obstacles (black) are static.
    The area enclosed in the rectangle with dashed boundaries are where most collisions occurred (depicted in detail in Fig. S2). 
}
    \label{fig:mbcx-steerbias}
\end{figure}

In the specific episode shown in Fig. \ref{fig:mbcx-steerbias}, simulating a robot with severe intrinsic leftwards bias \(b_\omega=0.5\) and a mild noise \(\sigma_\omega=0.05\),
the model found quickly a nearly optimal, mostly collision-free path from its collision experience in a handful of the early trials,
In contrast, without learning A2 always ran into the same obstacle(s) across all the trials,
as any differences in its paths were merely caused by the steering noise.

\begin{figure*}[ht!]
    \centering
    \includegraphics[width=\textwidth]{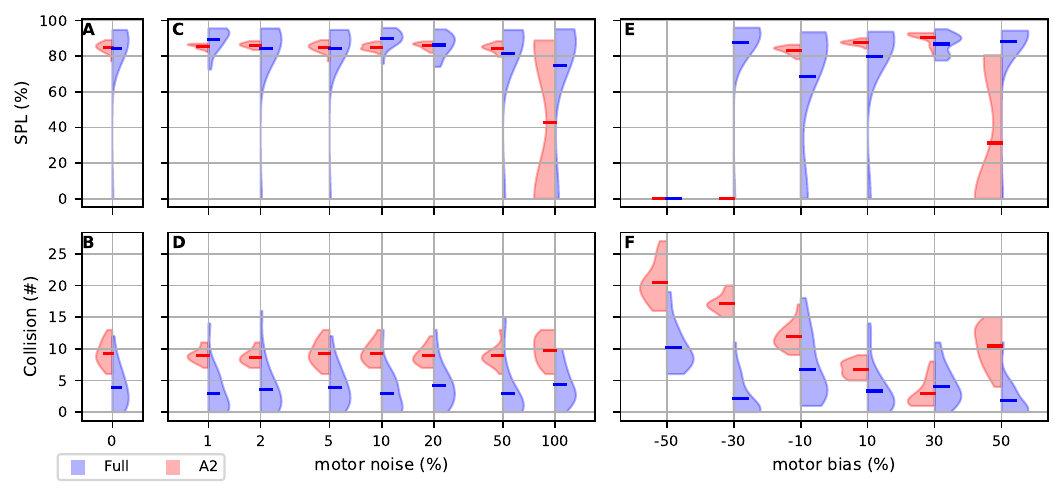}
    \caption{
    The effect of motor noise (\(\sigma_\omega\)) and bias (\(b_\omega\)) on performance. 
    \textbf{(AB)} No motor noise or bias (default by simulator).
    \textbf{(CD)} Additional motor noise only.
    \textbf{(EF)} Additional motor bias only.
    \(20\) consecutive trials were simulated for each scenario. 
    Note the trials were effectively mutually independent for A2,
    whereas their results were correlated for the full model (for learning).
    Details of trial-by-trial performance are shown in in Fig. S3 and S4.
    }
    \label{fig:gibson-noise-bias}
\end{figure*}

To examine the robustness of the full model in more detail, using the same episode, we varied \(\sigma_\omega\) and \(b_\omega\) separately (keeping the other at zero).
We found the magnitude of the steering noise (for \(\sigma_\omega \leq 1\)) has a relatively small impact on the performance of the full and ablated models (Fig. \ref{fig:gibson-noise-bias}A-D).
In general, the full model accrued significantly less collisions than A2 and almost always outperformed A2 in SPL, except for \(\sigma_\omega=0.5\) when the full model yielded a lower average SPL by approximately \(0.1\) (the median and mode were higher, though).
When \(\sigma_\omega=1\), A2 would fail to reach the goal about half of the times,
while the full model could still succeed consistently (Fig. \ref{fig:gibson-noise-bias}C).

The steering bias manifested a greater effect (Fig. \ref{fig:gibson-noise-bias}ABEF).
The performance of A2 was largely dependent on the bias.
A2 could not reach the goal completely when \(b_\omega \leq -0.3\) and more than half the times when \(b_\omega = 0.5\), 
as the rightwards bias drove the robot too south or north onto the west-facing wall (from \((-5, 0)\) to \((-5, 1)\) or north to \((-5, 2)\)),
preventing the robot from entering the west-to-east corridor.
A2's maximal performance was reached at \(b_\omega = 0.3\),
when paths were biased leftwards from the beeline into a shape
coincidentally similar to the last, optimised path of the full model in Fig. \ref{fig:mbcx-steerbias}.

In contrast, the full model was less affected by the bias.
Although it could not overcome the largest rightwards bias (\(b_\omega = -0.5\)),
failing to reach the goal due to the same obstacle as A2,
it succeeded consistently in all other cases (Fig. \ref{fig:gibson-noise-bias}E).
In particular, only the full model could adapt to \(b_\omega = -0.3\),
when A2 failed completely.

The robustness against steering noise can be explained by the same observation from our previous work on route following with the same robot in \igibson
that most noise effects were smoothed out by the realistic, low-pass filtering motors,
while the control commands were generated by the model at relatively high frequency \citep{yihe2025insect}.
It was even easier to mitigate the steering bias in point-goal navigation,
because the embodied model always knew the correct goal position.

Examining the trial-by-trial performance reveals that the full model indeed yielded less collisions and higher SPL most of the times,
while occasionally failing to reach the goal (\(\text{SPL}=0\) in Fig. S3 and S4).
Nevertheless, by rejecting the consolidation of the STM and ITM associated with the failures, the model could recover success immediately from the subsequent trial,
suggesting that the occasional failure was likely caused by randomness
and that the selective memory consolidation mechanism was effective in sustaining performance in non-deterministic environments.

\section{Discussion}
\label{sec-discussion}

Ants can learn to negotiate cluttered environments while using PI, i.e., dead reckoning, to relocate a target location such as their nest or a food source. 
In particular, after foraging and finding food, they attempt to return to their nest in a straight line, while quickly optimising their homing paths with respect to obstacles \citep{freas2025visual}.
This behaviour is formally equivalent to visual point-goal navigation, a popular benchmark in robotics. 
This motivates us to develop a novel model that integrates the MB and the CX,
two insect brain structures respectively responsible for associative learning and PI, and evaluate it for controlling a robot in point-goal navigation.
Compared to state-of-the-art RL or SLAM solutions for this task, our model is extremely lightweight while still capable of comparable performance, with zero training before testing.
It combines mapless motion planning with online learning in a minimal network with 64000 learnable parameters (more consistent with the brain size of ants and bees \citep{menzel2001cognitive, godfrey2021allometric}, compared to the millions used by the others (e.g., \citet{zhao2021surprising, partsey2022mapping}). 
The visual preprocessing is also much less expensive than the deep convolutional networks used in SLAM and RL (Table S2).

Similar insect-inspired learning circuits have been used in previous work, but we have introduced several key changes. First, most models of the visual learning circuit,
the MB, have focused on associating views to positive progress made continuously along a route \citep{ardin2016using, wystrach2020lateralised, gattaux2025route, yihe2025insect}, whereas our MB learns, in addition to attractive views just after escapes, repulsive views preceding a collision (which triggers reflexive escape).
Another difference is the introduction of selective memory consolidation, in the form of STM, ITM, and LTM processes consistent with biological learning systems \citep{davis2011traces}. 
These changes allow the MB to learn significantly fewer and less correlated views implies less interference in its memory and higher accuracy in obstacle recognition (\S\ref{sec-discussion:memory}).

Moreover, while integrative models of MB memory and CX performing PI have been studied before \citep{sun2020decentralised, goulard2023emergent},
they do not explain how learning and path finding affect each other.
Our model corroborates the hypothesis of \citet{clement2024latent} that an insect can `learn continuously the routes they travel'.

\subsection{Memory Interference and Consolidation}
\label{sec-discussion:memory}
While there are two MBONs in the MB neural network, learning occurs only in the KC-MBON layer and always at different times for the two MBONs.
The MB is thus equivalent to two separate networks with identical PN-KC weights but a single MBON,
each approximating a Bloom filter \citep{dasgupta2018neural}. 
\citet{ardin2016using} calculates the optimal memory capacity of such a single-MBON network explicitly,
\begin{equation}
\label{eq:mb-capacity}
    m = \ln\left(\frac{1-p_\text{error}^{1/N_\text{KC}}}{\kappa}\right) \bigg/\ln\left(1-\kappa\right),
\end{equation}
where \(\kappa=k/N_\text{KC}\) is the sparsity of KC activation
and \(p_\text{error}\) the likelihood of false positives, 
mistaking a novel input for a learnt one.
Different from their MB for route following, \(p_\text{error}\) here is the likelihood of our model perceiving obstacles while looking at unseen views.
We note, unlike many other neural network that may suffer from catastrophic forgetting \citep{french1999catastrophic}, it is impossible for a Bloom filer to `forget' old memories due to learning new ones and consequently generate a false negative response.
This means our model would never look at the same view of an obstacle that led to a previous collision without recognising and reacting to it (even though the reaction may not suffice successful obstacle avoidance).

\(p_\text{error}\) was extremely small in all our experiments,
because the maximal number of collisions in first trials of our model were less than \(100\) in all our experiments ('First' in Fig. \ref{fig:model-performance} and \ref{fig:gibson-noise-bias}),
implying the numbers of views learnt were always significantly less than \(100\) for each MBON, 
while a memory capacity of \(100\) views corresponds to \(p_\text{error}\approx 10^{-51}\) according to Eq. (\ref{eq:mb-capacity}).
Note this calculation is accurate only when every KC encoding of learnt views is independent of one another,
an optimal condition not guaranteed in practice.
Nevertheless, as our model is designed to learn visual features of obstacles that it has actually collided with, \(p_\text{error}\) is expected to be much higher for visually similar obstacles than traversable spaces.
This is desired to some extent, because the model is more likely to generalise its learnt avoidance behaviour correctly when facing obstacles.
However, this also risks over-generalisation.
The model may become over-cautious and `stuck' when facing obstacles, especially in a cluttered environment.

As shown in \S\ref{sec-results:memory-consolidation},
the adverse behavioural consequences of this over-generalisation risk was successfully mitigated by selective memory consolidation.
From the Bloom filter perspective, our implementation worked by managing the number of learnt views under the behaviourally (SPL) relevant memory capacity.
However, it sometimes led the model performance stuck in a local minima (\S\ref{sec-results:memory-consolidation}).
While this could be bypassed by steering noise (\S\ref{sec-result:robust-control}) or any other randomness,
it prompts consideration of alternative mechanisms for memory consolidation.

From the machine learning perspective, the trial-by-trial learning paradigm obeyed by our model can be formalised as a Markov Chain Monte Carlo process,
implying the possibility of applying corresponding methods, e.g., Metropolis–Hastings algorithm.
More specifically, Eq. (\ref{eq:ltm})-(\ref{eq:stm}) can be modified by adding gating rates for memory consolidation (rejection).

An alternative, more biomimetic approach may consider the memory acquisition and consolidation process time-dependent,
e.g., functionally following the temporal weighting rule as observed in ants \citep{freas2017learning}.
Such time-dependent mechanisms may also be more biologically realistic at the neural circuit level of MB,
in processing STM, ITM, and LTM, 
e.g., DPM modulation \citep{davis2011traces}, multi-MBON encoding \citep{gkanias2022incentive}, synaptic homeostasis in sleep \citep{bushey2011sleep, weiss2021sleep}.
Despite differences, both the temporal weighting rule and the synaptic homeostasis hypothesis assert memory decay, implying a preference of new memories over old ones.
This complements the current implementation and, if integrated appropriately, permits graceful forgetting in LTM,
which might prevent the model from becoming over-cautious due to overloaded memories.

\subsection{Learning with Sparse Rewards}
\label{sec-discussion:relation-to-RL}

As the MB learning is modulated by reinforcement signals, it might appear conceptually akin to RL,
and several recent works have described the learning mechanism in the MB as prediction error minimisation \citep{bennett2021learning, jurgensen2024prediction}, sometimes in an explicit RL framework \citep{lochner2024reinforcement}.
However, our model is distinct from standard RL.
The MB performs learning by the hetero-associative learning rule, rather than backpropagation.
The model does not attempt to minimise errors between its predictions and the true values of reinforcement signals via large-scale training,
because the model relies on the strong prior of goal approaching and only makes adaptive adjustments after collisions, i.e., deviation from the direction to the actual goal for obstacle avoidance. 
In RL jargon, our model exploits a default, non-random `policy' and improves it via online learning.
As this non-random policy yields sufficiently good performance,
our model is ready for deployment on any new platform in any novel environment.

Without such default policies, an RL model can rarely receive extrinsic reward in early stages of training,
as it is almost impossible to reach a goal with a purely random policy \citep{mishkin2019benchmarking, savva2019habitat}.
How to deal with such sparse rewards is a long-lasting challenge in RL in general,
and one common solution, employed by the state-of-the-art RL models for point-goal navigation, is reward engineering,
i.e., designing auxiliary, intermediate reinforcement signals that enable policy updating at a higher rate to accelerate RL. 
In particular, in addition to a large reward for reaching a goal, the RL models assume a small reward for moving closer to the goal.
Note `moving closer' is reward-engineered in terms of the geodesic distance, rather than Euclidean distance, to the goal in all these works \citep{savva2019habitat, wijmans2019dd, zhao2021surprising, partsey2022mapping},
effectively leaking (part of) the information of the shortest paths to the goal to the RL models.

We note that this choice of geodesic distance as an auxiliary reinforcement signal is not actually practical for a real robotic agent.
In a truly novel environment (with no prior mapping), accurate geodesic distance to the goal is always unknown.
If the geodesic distance could somehow be estimated without a map, 
it would be trivial for an agent to treat the geodesic information as a potential field and then solve point-goal navigation by gradient ascent (or descent),
making training obsolete. 
Various alternative choices of auxiliary reinforcement might be considered but any choice permits the risk of finding `reward hacking' solutions that deprioritise the actual goal,
e.g., abusing the sliding behaviour that occurs when colliding with obstacles in the simulation environment \citep{kadian2020sim2real}. 
To mitigate the risk, large-scale, parallel training is a necessity for the RL models to receive stable training signals \citep{wijmans2019dd}.

\subsection{Obstacle Detection and Avoidance}
\label{sec-discussion:obstacle-avoidacne}
Many existing insect-inspired models perform obstacle detection and avoidance by modelling insect vision with greater biological detail,
such as selecting the lowest point in skyline \citep{baddeley2012model,amin2025ant} or avoiding directions of looming \citep{sun2023insect}.
Alternatively, obstacle detection may be achieved by depth sensing.
Insects can estimate distances to objects by antennae and eyes,
and robots by LiDAR and depth camera.
On the same robot in \igibson, our previous work used depth data to reduce collisions \citep{yihe2025insect},
and performance gain in Habitat point-goal navigation was reported for all the models cited in Table \ref{tab:model-types} with RGBD inputs, compared to the same models but with RGB vision.

All these obstacle detection mechanisms are hardwired and thus immediately effective in testing.
In contrast, our model is built on the weaker assumption that visual detection of obstacles is to be learnt from scratch only after detection of collisions.
Consequently, obstacle detection and avoidance are completely adaptive in our model.
Despite the differences, it is viable to integrate our model with any hardwired mechanism, if only an appropriate temporary target direction can be set accordingly.

Such integration is expect to benefit the performance.
Effective obstacle avoidance is generally desired to reduce collisions,
which minimises the risk of physical damage for a real insect or robot.
Selecting lowest point in skyline or following obstacle boundary (as in bug algorithms \S\ref{sec-discussion:mapless-nav}) based on depth sensing
can be particularly beneficial to our model, especially in an environment cluttered with obstacles,
because they would guarantee the model facing and learning a view of no obstacles after escape,
whereas the current escape strategy may drive the robot into other obstacles.

\subsection{Navigation without a Map}
\label{sec-discussion:mapless-nav}
To navigate towards a goal without a map or mapping,
our model combines goal approaching and obstacle avoidance,
a strategy adopted by many mapless navigation models, including insect-inspired ones  \citep{baddeley2012model,sun2023insect,amin2025ant}.
As in all these insect-inspired models, goal approaching is manifested by the CX via PI (highly abstracted in our model).
Differently, instead of using a hardwired mechanism, our model learns obstacle avoidance from collision-triggered learning (\S\ref{sec-discussion:obstacle-avoidacne}).


Canonical examples of adaptive mapless navigation by default goal approaching and collision-triggered learning are bug algorithms \citep{mcguire2019comparative}.
In bug algorithms, it is assumed that boundaries of obstacles can be followed perfectly after collisions, and learning takes the form of explicitly and incrementally memorising an increasing number of collision coordinates. Neither of these occur in our model, and consequently there is no guarantee (as promised by bug algorithms) that our model would learn to avoid the collision positions or to reach a goal in an environment.
However, as our model learns to recognise obstacles visually, it may generalise learnt obstacle avoidance to novel or dynamic obstacles with similar appearances (\S\ref{sec-discussion:memory}),
whereas such generalisation is impossible in bug algorithms.
More crucially, in a realistic environment with odometry drifts and errors, bug algorithms are unable to uphold the theoretical guarantee of goal reaching, 
and, as the magnitude of (simulated) odometry drifts increases, the simplest bug algorithm which starts out as the worst ends up the best amongst all the bug algorithms tested in \citet{mcguire2019comparative}.

In VO+RL models \citep{zhao2021surprising, partsey2022mapping}, the problem of odometry drift is addressed by training the sensing-to-odometry and odometry-to-control pathways, allowing visual features to stabilise the estimate of self motion, position, and orientation. 
In insects, the robustness of odometry is significantly enhanced by the use of external compass cues, including visual information, with some ability to learn the relative stability of alternative cues  \citep{pahl2011large, wystrach2012ants}. 
The abstracted CX in our model should be replaced by one of such more realistic mechanisms, depending on specific applications.

\subsection{Choice of Environment and Baseline}
\label{sec-discussion:baseline}

Although our model yielded `similar' performance in the two simulators (in terms of SR and SPL),
the ablation study and the analysis on `hard' episodes show that the performance, using the same metrics, was significantly dependent on simulators and episodes.
In particular, our \igibson tasks were far easier than the \habitat ones (Fig. S1),
which might be rooted from differences in the low-level motion control, the particular indoor scenes, the objects contained in the scenes, or any other nuances.
Similarly, \citet{savva2019habitat} found model performance in \gibson 4+ scenes always better than that in MP3D, regardless of the training scenes in \gibson 4+ or MP3D scenes.

Indeed, all these scenes are probably designed to be easy for navigation, especially to humans,
as they are (primarily) civil facilities,
`including households, offices, hotels, venues, museums, hospitals, construction sites' \citep{xiazamirhe2018gibsonenv},
rather than prisons or military structures,
which may be designed to challenge or restrict navigation.
\citet{mishkin2019benchmarking} reports a human baseline with MP3D scenes,
scoring \(\text{SR}=0.953\) and \(\text{SPL}=0.867\) (with the main caveat that there was only \textit{one} participant).
The inherent traversability of such scenes may explain the non-zero performance of the non-learning, heuristic models (including A1 and A2 here).
Natural habitats of insects, especially to those capable of flying, are even more traversable than the civil facilities,
which may explain the strong reliance on PI in insect navigation.

For fairer benchmarking, it is useful to establish a `difficulty' metric for different episodes in any simulated or real environments. 
We suggest the performance of an odometry-only baseline might be more suitable metric than the current use of the ratio between the Euclidean and geodesic distances to the goal (termed `complexity' in Habitat datasets) \citep{savva2019habitat}.   
The odometry-only performance is straightforward to obtain 
and does not require full floor plans (for computing the geodesic distance).
It also naturally encompasses behaviourally relevant challenges in navigation, such as the higher difficulty to negotiate a small concave obstacle than a large convex one, which is not reflected in the conventional `complexity' metric.
More generally, the difficulty of point-navigation task is fundamentally determined by all potential interactions between a robot and its environment,
e.g., uncertainties in sensors and actuators, consequences of collisions (sliding or not).
Such interactions are affected by, but not reflected in, the static floor plan of a scene,
but can all be accounted for by the odometry-only baseline.

\subsection{Limitations and Future Work}
\label{sec-discussion:limitation}

\subsubsection{Testing on Real Robots}
The main limitation of this work is that our model was tested only in simulations, 
and, except when artificial noise was added,
the physics, the visual rendering, and the odometry data were unrealistically deterministic and consistent within and across tests.
While it is undoubtedly impressive that \citet{ramakrishnan2021habitat} managed to reach perfect performance in the Habitat task,
the authors admitted in Appendix 8 that their analysis on model performance `\textit{not} very indicative of the transfer performance to a real robot'.
Indeed, \citet{partsey2022mapping} tested their best model (V2 in Table \ref{tab:model-types}) in the real world without adaptation
and found the SR dropped from \(0.96\) to \(0.11\).

While our model faces a similar sim-to-real gap,
the gap is expect to be smaller for two reasons.
Firstly, our model does not require training before testing, and thus the size or the quality of training datasets does not matter.
In addition to the amount of training, the carefully curated training datasets made a noticeable contribution to training the near-perfect agents in \citet{wijmans2019dd, ramakrishnan2021habitat}.
Secondly, our model performs online learning in time limited trials,
during which environmental factors (e.g., ambient light) should change much less than across the transfer from offline training to testing afterwards, especially in cross-platform, novel scene scenarios.

\subsubsection{Optimisation} 
In addition, our model has not been optimised in parameter choice or algorithmic implementation or for hardware.
The user-defined parameters were chosen arbitrarily without optimisation (Table S1). 
In particular, the learning rate \(\alpha\) was set to be the highest possible learning rate without any tuning.
As the PN-KC weights and the KC activity are sparse and binary, in principle all computations in the MB can be further optimised for greater efficiency by executing addition (and subtraction) only.
The visual preprocessing (\S\ref{sec-method:visual-preprocessing}) is obsolete, if a low-resolution, monochrome camera can be used instead.

\subsubsection{Extending the MB Memory}
To improve the range and robustness of navigation,
our model can be extended by merging with other insect-inspired models capable of visual route following and homing.
This extension is readily viable, because the neural architectures of the MB are almost identical.
It is also more consistent with the real insect MB to have separate MBONs for learning attractive and repulsive stimuli \citep{aso2014mushroom}.
In particular, the homing behaviour is more robust with different MBONs learning views towards the goal and those in the opposite direction \citep{le2020opponent},
the efficacy of which has been demonstrated on a real robot by \citet{gattaux2025route}.
Functionally, this extension will result in an adaptive, visual potential field as used in local path planning.
To balance exploration and exploitation, the model should prioritise repulsive memories first and eventually rely more on attractive memories.

In addition, it is in principle viable to further mitigate drifting errors by adding more MBONs,
each responsible for learning and recognising views near different positions (landmarks),
effectively making the model SLAM-like by scaling up visual homing.

\section*{Supplementary Materials}
A supplementary containing Table S1 and S2 and Fig. S1-S4 can be found at \url{https://github.com/InsectRobotics/insect-inspired-point-goal-navigation/blob/main/pointgoal_paper_supplementary.pdf}.

All data were generated from simulation. 
The code for the simulation and the analysis are available at \url{https://github.com/InsectRobotics/insect-inspired-point-goal-navigation}.

\section*{Acknowledgments}
While LY is a current employee of the Universify of Glasgow, this research has been entirely carried out at the University of Edinburgh,
with the initial funding secured by BW from Huawei Technologies Co.,Ltd. [grant number YBN2020045132].

\bibliographystyle{elsarticle-harv} 
\bibliography{references}

@article{yihe2025insect,
  title={Insect-inspired embodied visual route following},
  author={Lu, Yihe and Cen, Jiahao and Alkhoury Maroun, Rana and Webb, Barbara},
  journal={Journal of Bionic Engineering},
  volume={22},
  number={3},
  pages={1167--1193},
  year={2025},
  publisher={Springer}
}

@article{dasgupta2018neural,
  title={A neural data structure for novelty detection},
  author={Dasgupta, Sanjoy and Sheehan, Timothy C and Stevens, Charles F and Navlakha, Saket},
  journal={Proceedings of the National Academy of Sciences},
  volume={115},
  number={51},
  pages={13093--13098},
  year={2018},
  publisher={National Acad Sciences},
  doi={10.1073/pnas.1814448115}
}

@InProceedings{li-iGibson,
  title = 	 {{iGibson} 2.0: object-centric Simulation for Robot Learning of Everyday Household Tasks},
  author =       {Li, Chengshu and Xia, Fei and Mart\'in-Mart\'in, Roberto and Lingelbach, Michael and Srivastava, Sanjana and Shen, Bokui and Vainio, Kent Elliott and Gokmen, Cem and Dharan, Gokul and Jain, Tanish and Kurenkov, Andrey and Liu, Karen and Gweon, Hyowon and Wu, Jiajun and Fei-Fei, Li and Savarese, Silvio},
  booktitle = 	 {Proceedings of the 5th Conference on Robot Learning},
  pages = 	 {455--465},
  year = 	 {2022},
  volume = 	 {164},
  series = 	 {Proceedings of Machine Learning Research},
  month = 	 {08--11 Nov},
  publisher =    {PMLR},
  url = 	 {https://proceedings.mlr.press/v164/li22b.html}
}

@article{ardin2016using,
  title={Using an insect mushroom body circuit to encode route memory in complex natural environments},
  author={Ardin, Paul and Peng, Fei and Mangan, Michael and Lagogiannis, Konstantinos and Webb, Barbara},
  journal={PLOS Computational Biology},
  volume={12},
  number={2},
  pages={e1004683},
  year={2016},
  publisher={Public Library of Science San Francisco, CA USA},
  doi={10.1371/journal.pcbi.1004683}
}

@article {wystrach2020lateralised,
  author = {Wystrach, Antoine and Le Mo{\"e}l, Florent and Clement, Leo and Schwarz, Sebastian},
  title = {A lateralised design for the interaction of visual memories and heading representations in navigating ants},
  elocation-id = {2020.08.13.249193},
  year = {2020},
  doi = {10.1101/2020.08.13.249193},
  publisher = {Cold Spring Harbor Laboratory},
  journal = {bioRxiv}
}

@article{steinbeck2024familiarity,
  author = {Steinbeck, Fabian and Kagioulis, Efstathios and Dewar, Alex and Philippides, Andrew and Nowotny, Thomas and Graham, Paul},
  title = {Familiarity-taxis: a bilateral approach to view-based snapshot navigation},
  year = {2024},
  issue_date = {Oct 2024},
  publisher = {Sage Publications, Inc.},
  address = {USA},
  volume = {32},
  number = {5},
  issn = {1059-7123},
  url = {https://doi.org/10.1177/10597123231221312},
  doi = {10.1177/10597123231221312},
  journal = {Adaptive Behavior},
  month = oct,
  pages = {407--420},
  numpages = {14}
}

@article{wystrach2023neurons,
  title={Neurons from pre-motor areas to the Mushroom bodies can orchestrate latent visual learning in navigating insects},
  author={Wystrach, Antoine},
  journal={bioRxiv},
  year={2023},
  publisher={Cold Spring Harbor Laboratory},
  doi={10.1101/2023.03.09.531867}
}

@article{sun2023insect,
  title={An insect-inspired model facilitating autonomous navigation by incorporating goal approaching and collision avoidance},
  author={Sun, Xuelong and Fu, Qinbing and Peng, Jigen and Yue, Shigang},
  journal={Neural Networks},
  volume={165},
  pages={106--118},
  year={2023},
  publisher={Elsevier},
  doi={10.1016/j.neunet.2023.05.033}
}

@article{webb2019internal,
  title={The internal maps of insects},
  author={Webb, Barbara},
  journal={Journal of Experimental Biology},
  volume={222},
  number={Suppl\_1},
  pages={jeb188094},
  year={2019},
  publisher={The Company of Biologists Ltd},
  doi={10.1242/jeb.188094}
}

@article{stone2017anatomically,
  title={An anatomically constrained model for path integration in the bee brain},
  author={Stone, Thomas and Webb, Barbara and Adden, Andrea and Weddig, Nicolai Ben and Honkanen, Anna and Templin, Rachel and Wcislo, William and Scimeca, Luca and Warrant, Eric and Heinze, Stanley},
  journal={Current Biology},
  volume={27},
  number={20},
  pages={3069--3085},
  year={2017},
  publisher={Elsevier},
  doi={10.1016/j.cub.2017.08.052}
}

@article{mussells2024converting,
  title={Converting an allocentric goal into an egocentric steering signal},
  author={Mussells Pires, Peter and Zhang, Lingwei and Parache, Victoria and Abbott, LF and Maimon, Gaby},
  journal={Nature},
  volume={626},
  number={8000},
  pages={808--818},
  year={2024},
  publisher={Nature Publishing Group UK London}
}

@article{mitchell2026framework,
  title={A framework for constructing insect steering circuits},
  author={Mitchell, Robert and Webb, Barbara},
  journal={PLOS Computational Biology},
  volume={22},
  number={4},
  pages={e1014009},
  year={2026},
  publisher={Public Library of Science San Francisco, CA USA}
}

@article{li2020connectome,
  title={The connectome of the adult \emph{Drosophila} mushroom body provides insights into function},
  author={Li, Feng and Lindsey, Jack W and Marin, Elizabeth C and Otto, Nils and Dreher, Marisa and Dempsey, Georgia and Stark, Ildiko and Bates, Alexander S and Pleijzier, Markus William and Schlegel, Philipp and others},
  journal={eLife},
  volume={9},
  pages={e62576},
  year={2020},
  publisher={eLife Sciences Publications, Ltd},
  doi={10.7554/eLife.62576}
}

@InProceedings{kodzhabashev2015route,
  author={Kodzhabashev, Aleksandar and Mangan, Michael},
  editor={Wilson, Stuart P. and Verschure, Paul F.M.J. and Mura, Anna and Prescott, Tony J.},
  title={Route Following Without Scanning},
  booktitle={Biomimetic and Biohybrid Systems.},
  year={2015},
  publisher={Springer International Publishing},
  address={Cham},
  pages={199--210},
  doi={10.1007/978-3-319-22979-9_20}
}

@article{schwarz2011properties,
  title={The properties of the visual system in the {Australian} desert ant \emph{Melophorus bagoti}},
  author={Schwarz, Sebastian and Narendra, Ajay and Zeil, Jochen},
  journal={Arthropod Structure \& Development},
  volume={40},
  number={2},
  pages={128--134},
  year={2011},
  publisher={Elsevier},
  doi={10.1016/j.asd.2010.10.003}
}

@article{hattori2017representations,
  title={Representations of novelty and familiarity in a mushroom body compartment},
  author={Hattori, Daisuke and Aso, Yoshinori and Swartz, Kurtis J and Rubin, Gerald M and Abbott, L F and Axel, Richard},
  journal={Cell},
  volume={169},
  number={5},
  pages={956--969},
  year={2017},
  publisher={Elsevier},
  doi={10.1016/j.cell.2017.04.028}
}

@article{wystrach2016field,
  title={How do field of view and resolution affect the information content of panoramic scenes for visual navigation? {A} computational investigation},
  author={Wystrach, Antoine and Dewar, Alex and Philippides, Andrew and Graham, Paul},
  journal={Journal of Comparative Physiology A},
  volume={202},
  pages={87--95},
  year={2016},
  publisher={Springer},
  doi={10.1007/s00359-015-1052-1}
}

@article{collett1992visual,
  title={Visual landmarks and route following in desert ants},
  author={Collett, Thomas S and Dillmann, Elisabeth and Giger, A and Wehner, R{\"u}diger},
  journal={Journal of Comparative Physiology A},
  volume={170},
  pages={435--442},
  year={1992},
  publisher={Springer},
  doi={10.1007/BF00191460}
}

@article{wystrach2011views,
  title={Views, landmarks, and routes: how do desert ants negotiate an obstacle course?},
  author={Wystrach, Antoine and Schwarz, Sebastian and Schultheiss, Patrick and Beugnon, Guy and Cheng, Ken},
  journal={Journal of Comparative Physiology A},
  volume={197},
  pages={167--179},
  year={2011},
  publisher={Springer},
  doi={10.1007/s00359-010-0597-2}
}

@inproceedings{zhao2021surprising,
  title={The surprising effectiveness of visual odometry techniques for embodied pointgoal navigation},
  author={Zhao, Xiaoming and Agrawal, Harsh and Batra, Dhruv and Schwing, Alexander},
  booktitle={2021 IEEE/CVF International Conference on Computer Vision (ICCV)},
  pages={16107--16116},
  year={2021},
  doi={10.1109/ICCV48922.2021.01582}
}

@INPROCEEDINGS{partsey2022mapping,
  author={Partsey, Ruslan and Wijmans, Erik and Yokoyama, Naoki and Dobosevych, Oles and Batra, Dhruv and Maksymets, Oleksandr},
  booktitle={2022 IEEE/CVF Conference on Computer Vision and Pattern Recognition (CVPR)},
  title={Is Mapping Necessary for Realistic PointGoal Navigation?},
  year={2022},
  pages={17211--17220},
  doi={10.1109/CVPR52688.2022.01672}
}

@article{labrosse2006visual,
  title={The visual compass: Performance and limitations of an appearance-based method},
  author={Labrosse, Fr{\'e}d{\'e}ric},
  journal={Journal of Field Robotics},
  volume={23},
  number={10},
  pages={913--941},
  year={2006},
  publisher={Wiley Online Library},
  doi={10.1002/rob.20159}
}

@article{hulse2021connectome,
  title={A connectome of the \emph{Drosophila} central complex reveals network motifs suitable for flexible navigation and context-dependent action selection},
  author={Hulse, Brad K and Haberkern, Hannah and Franconville, Romain and Turner-Evans, Daniel and Takemura, Shin-ya and Wolff, Tanya and Noorman, Marcella and Dreher, Marisa and Dan, Chuntao and Parekh, Ruchi and others},
  journal={eLife},
  volume={10},
  year={2021},
  publisher={eLife Sciences Publications, Ltd},
  doi={10.7554/eLife.66039}
}

@article{philippides2011might,
  title={How might ants use panoramic views for route navigation?},
  author={Philippides, Andrew and Baddeley, Bart and Cheng, Ken and Graham, Paul},
  journal={Journal of Experimental Biology},
  volume={214},
  number={3},
  pages={445--451},
  year={2011},
  publisher={Company of Biologists},
  doi={10.1242/jeb.046755}
}

@article{baddeley2012model,
  title={A model of ant route navigation driven by scene familiarity},
  author={Baddeley, Bart and Graham, Paul and Husbands, Philip and Philippides, Andrew},
  journal={PLOS Computational Biology},
  volume={8},
  number={1},
  pages={e1002336},
  year={2012},
  publisher={Public Library of Science San Francisco, USA},
  doi={10.1371/journal.pcbi.1002336}
}

@article{mcguire2019comparative,
  title={A comparative study of bug algorithms for robot navigation},
  author={McGuire, Kimberly N and de Croon, Guido C H E and Tuyls, Karl},
  journal={Robotics and Autonomous Systems},
  volume={121},
  pages={103261},
  year={2019},
  publisher={Elsevier},
  doi={10.1016/j.robot.2019.103261}
}

@article{goulard2023emergent,
  title={Emergent spatial goals in an integrative model of the insect central complex},
  author={Goulard, Roman and Heinze, Stanley and Webb, Barbara},
  journal={PLOS Computational Biology},
  volume={19},
  number={12},
  pages={e1011480},
  year={2023},
  publisher={Public Library of Science San Francisco, CA USA},
  doi={10.1371/journal.pcbi.1011480}
}

@article{gattaux2025route,
  title={Route-centric ant-inspired memories enable panoramic route-following in a car-like robot},
  author={Gattaux, Gabriel G and Wystrach, Antoine and Serres, Julien R and Ruffier, Franck},
  journal={Nature Communications},
  volume={16},
  number={1},
  pages={8328},
  year={2025},
  publisher={Nature Publishing Group UK London}
}

@article{mishkin2019benchmarking,
  title={Benchmarking classic and learned navigation in complex {3D} environments},
  author={Mishkin, Dmytro and Dosovitskiy, Alexey and Koltun, Vladlen},
  journal={arXiv preprint arXiv:1901.10915},
  year={2019}
}

@inproceedings{savva2019habitat,
  title={Habitat: a platform for embodied {AI} research},
  author={Savva, Manolis and Kadian, Abhishek and Maksymets, Oleksandr and Zhao, Yili and Wijmans, Erik and Jain, Bhavana and Straub, Julian and Liu, Jia and Koltun, Vladlen and Malik, Jitendra and others},
  booktitle={Proceedings of the IEEE/CVF International Conference on Computer Vision},
  pages={9339--9347},
  year={2019}
}

@article{wijmans2019dd,
  title={{DD-PPO}: learning near-perfect pointgoal navigators from 2.5 billion frames},
  author={Wijmans, Erik and Kadian, Abhishek and Morcos, Ari and Lee, Stefan and Essa, Irfan and Parikh, Devi and Savva, Manolis and Batra, Dhruv},
  journal={arXiv preprint arXiv:1911.00357},
  year={2019}
}

@inproceedings{xiazamirhe2018gibsonenv,
  title={Gibson {Env}: real-world perception for embodied agents},
  author={Xia, Fei and R. Zamir, Amir and He, Zhi-Yang and Sax, Alexander and Malik, Jitendra and Savarese, Silvio},
  booktitle={Proceedings of the IEEE Conference on Computer Vision and Pattern Recognition},
  pages={9068-9079},
  year={2018},
  organization={IEEE}
}

@article{menzel2001cognitive,
  title={Cognitive architecture of a mini-brain: the honeybee},
  author={Menzel, Randolf and Giurfa, Martin},
  journal={Trends in Cognitive Sciences},
  volume={5},
  number={2},
  pages={62--71},
  year={2001},
  publisher={Elsevier}
}

@article{godfrey2021allometric,
  title={Allometric analysis of brain cell number in {Hymenoptera} suggests ant brains diverge from general trends},
  author={Godfrey, R Keating and Swartzlander, Mira and Gronenberg, Wulfila},
  journal={Proceedings of the Royal Society B: Biological Sciences},
  volume={288},
  number={1947},
  pages={20210199},
  year={2021},
  publisher={The Royal Society}
}

@inproceedings{stankiewicz2020using,
  title={Using the neural circuit of the insect central complex for path integration on a micro aerial vehicle},
  author={Stankiewicz, Jan and Webb, Barbara},
  booktitle={Conference on Biomimetic and Biohybrid Systems},
  pages={325--337},
  year={2020},
  organization={Springer}
}

@article{sun2020decentralised,
  title={A decentralised neural model explaining optimal integration of navigational strategies in insects},
  author={Sun, Xuelong and Yue, Shigang and Mangan, Michael},
  journal={eLife},
  volume={9},
  pages={e54026},
  year={2020},
  publisher={eLife Sciences Publications, Ltd}
}

@inproceedings{zhong2025comparative,
  title={A Comparative Study of Reinforcement Learning and Insect-Inspired Visual Navigation Methods},
  author={Zhong, Xiaoting and Sun, Xuelong and Li, Haiyang},
  booktitle={Conference on Biomimetic and Biohybrid Systems},
  pages={163--175},
  year={2025},
  organization={Springer}
}

@article{le2020opponent,
  title={Opponent processes in visual memories: A model of attraction and repulsion in navigating insects’ mushroom bodies},
  author={Le M{\"o}el, Florent and Wystrach, Antoine},
  journal={PLOS Computational Biology},
  volume={16},
  number={2},
  pages={e1007631},
  year={2020},
  publisher={Public Library of Science San Francisco, CA USA}
}

@article{lochner2024reinforcement,
  title={Reinforcement learning as a robotics-inspired framework for insect navigation: from spatial representations to neural implementation},
  author={Lochner, Stephan and Honerkamp, Daniel and Valada, Abhinav and Straw, Andrew D},
  journal={Frontiers in Computational Neuroscience},
  volume={18},
  pages={1460006},
  year={2024},
  publisher={Frontiers Media SA}
}

@article{bennett2021learning,
  title={Learning with reinforcement prediction errors in a model of the \emph{Drosophila} mushroom body},
  author={Bennett, James EM and Philippides, Andrew and Nowotny, Thomas},
  journal={Nature Communications},
  volume={12},
  number={1},
  pages={2569},
  year={2021},
  publisher={Nature Publishing Group UK London}
}

@article{jurgensen2024prediction,
  title={Prediction error drives associative learning and conditioned behavior in a spiking model of \emph{Drosophila} larva},
  author={J{\"u}rgensen, Anna-Maria and Sakagiannis, Panagiotis and Schleyer, Michael and Gerber, Bertram and Nawrot, Martin Paul},
  journal={iScience},
  volume={27},
  number={1},
  year={2024},
  publisher={Elsevier}
}

@article{kadian2020sim2real,
  title={{Sim2real} predictivity: does evaluation in simulation predict real-world performance?},
  author={Kadian, Abhishek and Truong, Joanne and Gokaslan, Aaron and Clegg, Alexander and Wijmans, Erik and Lee, Stefan and Savva, Manolis and Chernova, Sonia and Batra, Dhruv},
  journal={IEEE Robotics and Automation Letters},
  volume={5},
  number={4},
  pages={6670--6677},
  year={2020},
  publisher={IEEE}
}

@article{savva2017minos,
  author = {Manolis Savva and Angel X. Chang and Alexey Dosovitskiy and Thomas Funkhouser and Vladlen Koltun},
  title = {{MINOS}: multimodal Indoor Simulator for Navigation in Complex Environments},
  journal = {arXiv preprint arXiv:1712.03931},
  year = {2017}
}

@article{ramakrishnan2021habitat,
  title={{Habitat-Matterport 3D dataset (HM3D): 1000 large-scale {3D} environments for embodied AI}},
  author={Ramakrishnan, Santhosh K and Gokaslan, Aaron and Wijmans, Erik and Maksymets, Oleksandr and Clegg, Alex and Turner, John and Undersander, Eric and Galuba, Wojciech and Westbury, Andrew and Chang, Angel X and others},
  journal={arXiv preprint arXiv:2109.08238},
  year={2021}
}

@article{marr1980theory,
  title={Theory of edge detection},
  author={Marr, David and Hildreth, Ellen},
  journal={Proceedings of the Royal Society of London. Series B. Biological Sciences},
  volume={207},
  number={1167},
  pages={187--217},
  year={1980},
  publisher={The Royal Society London}
}

@article{seung2024predicting,
  title={Predicting visual function by interpreting a neuronal wiring diagram},
  author={Seung, H Sebastian},
  journal={Nature},
  volume={634},
  number={8032},
  pages={113--123},
  year={2024},
  publisher={Nature Publishing Group UK London}
}

@article{davis2011traces,
  title={Traces of \emph{Drosophila} memory},
  author={Davis, Ronald L},
  journal={Neuron},
  volume={70},
  number={1},
  pages={8--19},
  year={2011},
  publisher={Elsevier}
}

@article{gkanias2022incentive,
  title={An incentive circuit for memory dynamics in the mushroom body of \emph{Drosophila melanogaster}},
  author={Gkanias, Evripidis and McCurdy, Li Yan and Nitabach, Michael N and Webb, Barbara},
  journal={eLife},
  volume={11},
  pages={e75611},
  year={2022},
  publisher={eLife Sciences Publications Limited}
}

@article{weiss2021sleep,
  title={Sleep deprivation results in diverse patterns of synaptic scaling across the \emph{Drosophila} mushroom bodies},
  author={Weiss, Jacqueline T and Donlea, Jeffrey M},
  journal={Current Biology},
  volume={31},
  number={15},
  pages={3248--3261},
  year={2021},
  publisher={Elsevier}
}

@article{goldschmidt2017neurocomputational,
  title={A neurocomputational model of goal-directed navigation in insect-inspired artificial agents},
  author={Goldschmidt, Dennis and Manoonpong, Poramate and Dasgupta, Sakyasingha},
  journal={Frontiers in Neurorobotics},
  volume={11},
  pages={20},
  year={2017},
  publisher={Frontiers Media SA}
}

@article{aso2014mushroom,
  title={Mushroom body output neurons encode valence and guide memory-based action selection in \emph{Drosophila}},
  author={Aso, Yoshinori and Sitaraman, Divya and Ichinose, Toshiharu and Kaun, Karla R and Vogt, Katrin and Belliart-Gu{\'e}rin, Ghislain and Pla{\c{c}}ais, Pierre-Yves and Robie, Alice A and Yamagata, Nobuhiro and Schnaitmann, Christopher and others},
  journal={eLife},
  volume={3},
  pages={e04580},
  year={2014},
  publisher={eLife Sciences Publications, Ltd}
}

@article{freas2025visual,
  title={Visual learning, route formation and the choreography of looking back in desert ants, \emph{Melophorus bagoti}},
  author={Freas, Cody A and Cheng, Ken},
  journal={Animal Behaviour},
  volume={222},
  pages={123125},
  year={2025},
  publisher={Elsevier}
}

@article{freas2017learning,
  title={Learning and time-dependent cue choice in the desert ant, \emph{Melophorus bagoti}},
  author={Freas, Cody A and Cheng, Ken},
  journal={Ethology},
  volume={123},
  number={8},
  pages={503--515},
  year={2017},
  publisher={Wiley Online Library}
}

@article{freas2019terrestrial,
  title={Terrestrial cue learning and retention during the outbound and inbound foraging trip in the desert ant, Cataglyphis velox},
  author={Freas, Cody A and Spetch, Marcia L},
  journal={Journal of Comparative Physiology A},
  volume={205},
  number={2},
  pages={177--189},
  year={2019},
  publisher={Springer}
}

@article{clement2024latent,
  title={Latent learning without map-like representation of space in navigating ants},
  author={Clement, Leo and Schwarz, Sebastian and Wystrach, Antoine},
  journal={bioRxiv},
  year={2024},
  publisher={Cold Spring Harbor Laboratory}
}

@article{muller1988path,
  title={Path integration in desert ants, \emph{Cataglyphis fortis}},
  author={M{\"u}ller, Martin and Wehner, R{\"u}diger},
  journal={Proceedings of the National Academy of Sciences},
  volume={85},
  number={14},
  pages={5287--5290},
  year={1988}
}

@article{french1999catastrophic,
  title={Catastrophic forgetting in connectionist networks},
  author={French, Robert M},
  journal={Trends in Cognitive Sciences},
  volume={3},
  number={4},
  pages={128--135},
  year={1999},
  publisher={Elsevier}
}

@article{aso2014neuronal,
  title={The neuronal architecture of the mushroom body provides a logic for associative learning},
  author={Aso, Yoshinori and Hattori, Daisuke and Yu, Yang and Johnston, Rebecca M and Iyer, Nirmala A and Ngo, Teri-TB and Dionne, Heather and Abbott, LF and Axel, Richard and Tanimoto, Hiromu and others},
  journal={eLife},
  volume={3},
  pages={e04577},
  year={2014},
  publisher={eLife Sciences Publications, Ltd}
}

@article{webb2024beyond,
  title={Beyond prediction error: 25 years of modeling the associations formed in the insect mushroom body},
  author={Webb, Barbara},
  journal={Learning \& Memory},
  volume={31},
  number={5},
  pages={a053824},
  year={2024},
  publisher={Cold Spring Harbor Lab}
}

@article{moller2000insect,
  title={Insect visual homing strategies in a robot with analog processing},
  author={M{\"o}ller, Ralf},
  journal={Biological Cybernetics},
  volume={83},
  number={3},
  pages={231--243},
  year={2000},
  publisher={Springer}
}

@article{wystrach2012ants,
  title={Ants might use different view-matching strategies on and off the route},
  author={Wystrach, Antoine and Beugnon, Guy and Cheng, Ken},
  journal={Journal of Experimental Biology},
  volume={215},
  number={1},
  pages={44--55},
  year={2012},
  publisher={Company of Biologists}
}

@article{pahl2011large,
  title={Large scale homing in honeybees},
  author={Pahl, Mario and Zhu, Hong and Tautz, J{\"u}rgen and Zhang, Shaowu},
  journal={PLOS One},
  volume={6},
  number={5},
  pages={e19669},
  year={2011},
  publisher={Public Library of Science San Francisco, USA}
}

@article{le2019central,
  title={The central complex as a potential substrate for vector based navigation},
  author={Le Mo{\"e}l, Florent and Stone, Thomas and Lihoreau, Mathieu and Wystrach, Antoine and Webb, Barbara},
  journal={Frontiers in Psychology},
  volume={10},
  pages={690},
  year={2019},
  publisher={Frontiers Media SA}
}

@article{porr2003isotropic,
  title={Isotropic-sequence-order learning in a closed-loop behavioural system},
  author={Porr, Bernd and W{\"o}rg{\"o}tter, Florentin},
  journal={Philosophical Transactions of the Royal Society of London. Series A: Mathematical, Physical and Engineering Sciences},
  volume={361},
  number={1811},
  pages={2225--2244},
  year={2003},
  publisher={The Royal Society}
}

@article{tanimoto2004event,
  title={Event timing turns punishment to reward},
  author={Tanimoto, Hiromu and Heisenberg, Martin and Gerber, Bertram},
  journal={Nature},
  volume={430},
  number={7003},
  pages={983--983},
  year={2004},
  publisher={Nature Publishing Group UK London}
}

@article{noorman2024maintaining,
  title={Maintaining and updating accurate internal representations of continuous variables with a handful of neurons},
  author={Noorman, Marcella and Hulse, Brad K and Jayaraman, Vivek and Romani, Sandro and Hermundstad, Ann M},
  journal={Nature Neuroscience},
  volume={27},
  number={11},
  pages={2207--2217},
  year={2024},
  publisher={Nature Publishing Group US New York}
}

@article{amin2025ant,
  title={Ant visual route navigation: How the fine details of behaviour promote successful route performance and convergence},
  author={Amin, Amany Azevedo and Philippides, Andrew and Graham, Paul},
  journal={PLOS Computational Biology},
  volume={21},
  number={9},
  pages={e1012798},
  year={2025},
  publisher={Public Library of Science San Francisco, CA USA}
}

@article{bushey2011sleep,
  title={Sleep and synaptic homeostasis: structural evidence in Drosophila},
  author={Bushey, Daniel and Tononi, Giulio and Cirelli, Chiara},
  journal={Science},
  volume={332},
  number={6037},
  pages={1576--1581},
  year={2011},
  publisher={American Association for the Advancement of Science}
}

\end{document}